\documentclass[runningheads]{llncs}
\usepackage{graphicx}
\usepackage{amsmath}
\usepackage{xcolor}
\usepackage{booktabs,amsfonts,dcolumn}
\usepackage{multirow}

\usepackage{linguex}

\usepackage{subfig}
\newcolumntype{d}[1]{D..{#1}}
\usepackage{booktabs}
\usepackage{times}
\usepackage{soul}
\usepackage{url}
\usepackage[hidelinks]{hyperref}
\usepackage[utf8]{inputenc}
\usepackage{amsmath}

\usepackage{latexsym} 
\usepackage{floatrow}

\newfloatcommand{capbtabbox}{table}[][\FBwidth]
\usepackage{booktabs}

\begin{document}

\title{Zero-Shot Recommendation as Language Modeling}

\author{Damien Sileo\orcidID{0000-0002-3274-291X} \and
Wout Vossen \and
Robbe Raymaekers}
\authorrunning{D. Sileo et al.}

\institute{KU Leuven, Belgium\\
\email{damien.sileo@kuleuven.be}
}
%\iffalse
%}
%\fi

%\author{Anonymous Author}

\maketitle

\begin{abstract}
Recommendation is the task of ranking items (e.g. movies or products) according to individual user needs. Current systems rely on collaborative filtering and content-based techniques, which both require structured training data. We propose a framework for recommendation with off-the-shelf pretrained language models (LM) that only used unstructured text corpora as training data. If a user $u$ liked \textit{Matrix} and \textit{Inception}, we construct a textual prompt, e.g. \textit{"Movies like Matrix, Inception, ${<}m{>}$"} to estimate the affinity between  $u$ and $m$ with LM likelihood. We motivate our idea with a corpus analysis, evaluate several prompt structures, and we compare LM-based recommendation with standard matrix factorization trained on different data regimes. The code for our experiments is publicly available\footnote{\href{https://colab.research.google.com/drive/1f1mlZ-FGaLGdo5rPzxf3vemKllbh2esT?usp=sharing}{\texttt{https://colab.research.google.com/drive/...?usp=sharing}}}.

\end{abstract}

\section{Introduction}
Recommender systems predict an affinity score between users and items. Current recommender systems are based on content-based filtering (CB),  collaborative filtering techniques (CF), or a combination of both.  CF recommender systems rely on (\textsc{User,  Item,  Interaction}) triplets.  CB relies on (\textsc{Item, Features}) pairs. Both system types require a costly structured data collection step.  Meanwhile, web users  express themselves about various items in an unstructured way. They share lists of their favorite items and ask for recommendations on web forums, as in \ref{ex1}\footnote{\href{https://www.reddit.com/r/MovieSuggestions/comments/cuuwrk/films_like_beyond_the_black_rainbow_lost_river/} {\texttt{https://www.reddit.com/r/MovieSuggestions/...lost\_river/}}} which hints at a similarity between the enumerated movies. 
\ex. \textit{\hspace{-0.3cm} Films like Beyond the Black Rainbow, Lost River, Suspiria, and The Neon Demon. \label{ex1}}

\vspace{-0.5cm}
The web also contains a lot of information about the items themselves, like synopsis or reviews for movies. Language models such as GPT-2 \cite{radford2019language} are trained on large web corpora to generate plausible text. We hypothesize that they can make use of this unstructured knowledge to make recommendations by estimating the plausibility of items being grouped together in a prompt. LM can estimate the probability of a word sequence, $P(w_1,...w_n)$.  Neural language models are trained over a large corpus of documents: to train a neural network, its parameters $\Theta$ are optimized for next word prediction likelihood maximization over $k$-length sequences sampled from a corpus. The loss writes as follows:
\begin{equation}
\mathcal{L}_\text{LM} = - \text{log } \sum\limits_{i} P (w_i|w_{i-k}....w_{i-1};\Theta) \label{eq:languagemodeling}
\end{equation}

We rely on existing pretrained language models. To make a relevance prediction , we build a prompt for each user:
\vspace{-0.2cm}
\begin{equation}
p_{u,i}= \textit{Movies like } {<}m_1{>},...{<}m_n{>}, {<}m_i{>} 
\end{equation}
where ${<}m_i{>}$ is the name of the movie $m_i$ and ${<}m_1...m_n{>}$ are those of randomly ordered movies liked by $u$. We then directly use $\widehat{R}_{u,i} =P_\Theta(p_{u,i})$ as a relevance score to sort items for user $u$.

Our contributions are as follow (i) we propose a model for recommendation with standard LM; (ii) we derive prompt structures from a corpus analysis and compare their impact on recommendation accuracy; (iii) we compare LM-based recommendation with next sentence prediction (NSP) \cite{Penha20} and a standard supervised matrix factorization method \cite{koren2009matrix,rendle2012bpr}.

\section{Related work}

\paragraph{\textbf{Language models and recommendation}}  
Previous work leveraged language modeling techniques to perform recommendations. However, they do not rely on natural language: they use sequences of user/item interactions, and treat these sequences as sentences to leverage the architectures inspired by NLP, such as Word2Vec \cite{guardia2015latent,barkan2017item2vec,devooght2017long,li2018learning} or BERT \cite{BERT4Recs}. %By contrast, our model is the first to our knowledge to evaluate unsupervised recommendation by using internet text as the only supervision.
\vspace{-0.15cm}

\paragraph{\textbf{Zero-shot prediction with language models}} Neural language models have been used for zero-shot inference on many NLP tasks \cite{radford2019language,brown2020language}. For example, they manually construct a prompt structure to translate text, e.g. \textit{Translate english to french : "cheese" =${>}$}, and use the language model completions to find the best translations.   Petroni et al. \cite{petroni2019} show that masked language models can act as a knowledge base when we use part of a triplet as input, e.g. \textit{Paris in ${<}mask{>}$}.  Here, we apply LM-based prompts to recommendation.
\vspace{-0.15cm}

\paragraph{\textbf{Hybrid and zero-shot recommendation}}  The cold start problem \cite{Schein2002}, i.e. dealing with new users or items is a long-standing problem in recommender systems, usually addressed with hybridization of CF-based and CB-based systems.  Previous work \cite{Volkovs17,li2019zero,ding2021zeroshot,Feng21} introduced models for zero-shot recommendation, but they use zero-shot prediction with a different sense than ours. They train on a set of  {\textsc{(User, Item, Interaction)}} triplets, and perform zero-shot predictions on new users or items with known attributes.
These methods still require \textsc{(User, Item, Interaction)} or \textsc{(Item, Features)} tuples for training. To our knowledge, the only attempt to perform recommendations without such data at all is from Penha et al. \cite{Penha20} who showed that BERT \cite{devlin-etal-2019-bert} next sentence prediction (NSP) can be used to predict the most plausible movie after a prompt. NSP is not available in all language models and requires a specific pretraining. Their work is designed as a probing of BERT knowledge about common items, and lacks comparison with a standard recommendation model, which we here address.

%\cite{li2019zero}

\section{Experiments}
\subsection{Setup}
 \vspace{-0.1cm}

\paragraph{\textbf{Dataset}} We use the standard the MovieLens 1M dataset \cite{harper15} with 1M ratings from $0.5$ to $5$, 6040 users, and 3090 movies in our experiments. We address the relevance prediction task\footnote{Item relevance could be mapped to ratings but we do not address rating prediction here.}, so we consider a rating $r$ as positive if $r \geq 4.0$, as negative if  $\leq 2.5$ and we discard the other ratings. We select users with at least $21$ positive ratings and $4$ negative ratings and thus obtain $2716$ users. We  randomly select $20\%$ of them as test users\footnote{Training users are only used for the matrix factorization baseline.}. $1$ positive and $4$ negative ratings are reserved for evaluation for each user, and the goal is to give the highest relevance score to the positively rated item. We use $5$ positive ratings per user unless mentioned otherwise. We remove the years from the movie titles and reorder the articles (\textit{a, the}) in the movie titles provided in the dataset (e.g. \textit{Matrix, The (1999)} $\rightarrow$ \textit{The Matrix}).
 \vspace{-0.25cm}

 \paragraph{\textbf{Evaluation metric}} We use the mean average precision at rank 1 (MAP@1) \cite{schroder2011setting} which is the rate of correct first ranked prediction averaged over test users, because of its interpretability.
 \vspace{-0.25cm}
 
\paragraph{\textbf{Pretrained language models}} In our experiments we use the GPT-2 \cite{radford2019language} language models, which are publicly available in several sizes. GPT-2 is trained with LM pretraining (equation \ref{eq:languagemodeling}) on the WebText corpus \cite{radford2019language}, which contains 8 million pages covering various domains. Unless mentioned otherwise, we use the GPT-base model, with 117M parameters.

 \vspace{-0.1cm}

\subsection{Mining prompts for recommendation}
 \vspace{-0.1cm}

\ffigbox[\textwidth]{

    \begin{floatrow}

    \capbtabbox[0.99\linewidth]{
    \begin{tabular}{lllll}
    \cline{1-2}
    \multicolumn{1}{l}{\textbf{3-6 gram}} & \multicolumn{1}{l}{\textbf{\#Count}}  \\ \cline{1-2}
    
    ${<}m{>}$ \textit{and} ${<}m{>}$                                     & 387                                                                      \\
    ${<}m{>}$, ${<}m{>}$, ${<}m{>}$                           & 232                                                                 \\
    \textit{Movies like} ${<}m{>}$                                         & 196                                                                  \\
    ${<}m{>}$, ${<}m{>}$, ${<}m{>}$, ${<}m{>}$    \hspace{-0.5cm}         & 85                                                                       \\
    \textit{Movies similar to} ${<}m{>}$                                   & 25                                                                    \\
    \cline{1-2}    
    \end{tabular}
    \vspace{0.75cm}
    \caption{Occurrence counts of 3-6 grams that contain movie names in the Reddit corpus. ${<}m{>}$ denotes a movie name.} 
    \label{queryanalyse}
    }

    \ffigbox[0.99\linewidth]{
     \hspace*{-1cm}
        \includegraphics[width=1.0\columnwidth]{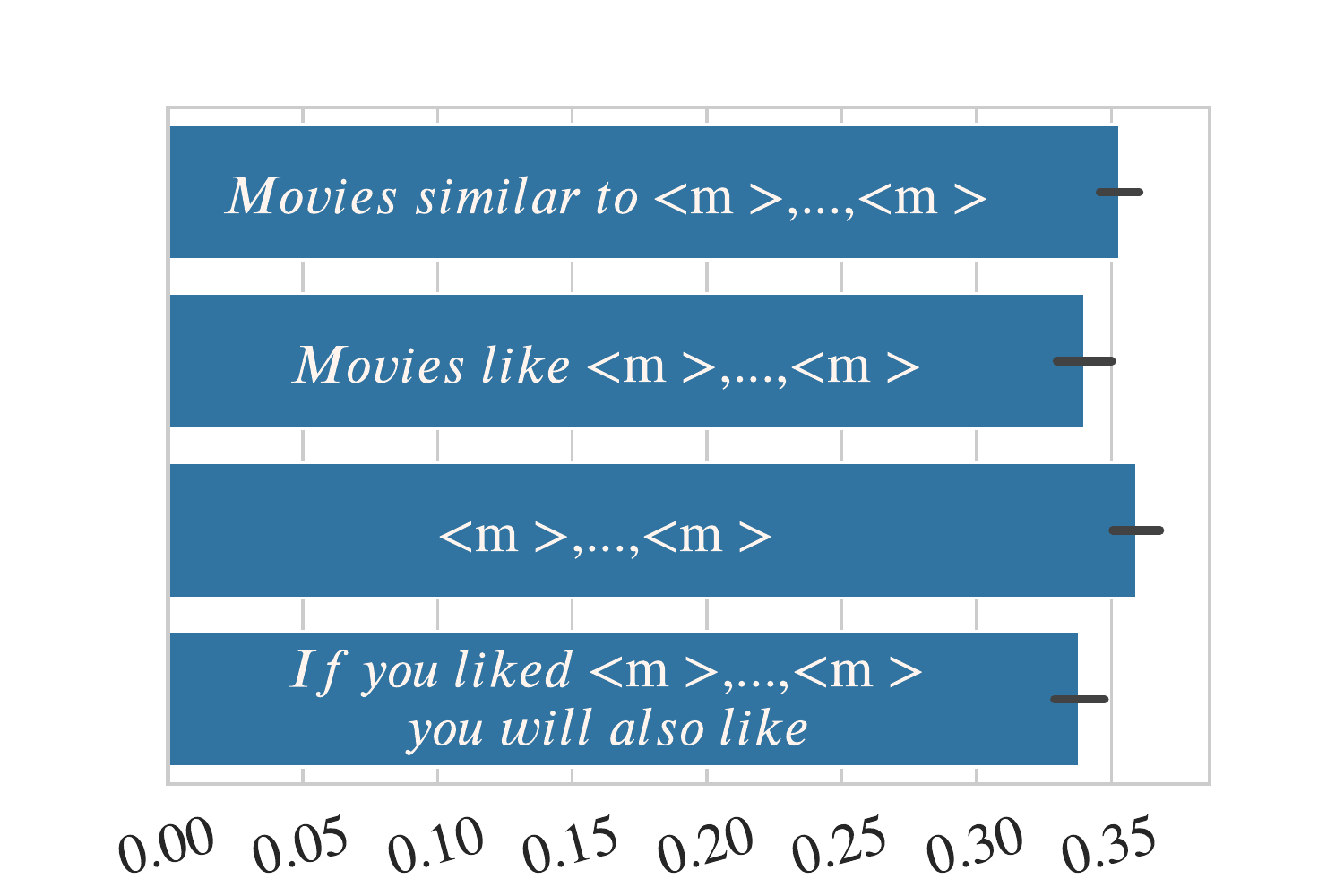}
        \caption{Comparison of LM recommendations MAP@1 with different prompt structures.\\}
     \label{fig:prompts}
    }
    \
    \end{floatrow}
}

 We analyze the Reddit comments from May 2015\footnote{\url{https://www.kaggle.com/reddit/reddit-comments-may-2015}} to find out how web users mention lists of movies in web text. This analysis will provide prompt candidates for LM-based recommendations.  We select comments where a movie name of the MovieLens dataset is present and replace movies with a ${<}m{>}$ tag.  This filtered dataset of comments has a size of $>900k$ words. We then select the most frequent pattern with at least three words, as shown in table \ref{queryanalyse}. Movie names are frequently used in enumerations. The patterns \textit{Movies like ${<}m{>}$} and \textit{Movies similar to } confirm that users focus on the similarity of movies.

Figure \ref{fig:prompts} shows that prompt design is important but not critical for high accuracy. Our corpus-derived prompts significantly outperform \textit{if you like ${<}m_1...m_n{>}$, you will like ${<}m_i{>}$} used in \cite{Penha20}. We will use ${<}m_1...m_n{>},{<}m_i{>}$ in the remaining of the paper due to its superior results and its simplicity.

\subsection{Effect of the number of ratings per test user}
We investigate the effect of the number of mentioned movies in prompts. We expect the accuracy of the models in making recommendations to increase when they get more info about movies a user likes. We compare the recommendation accuracy on the same users 0,1,2,3,5,10,15 or 20 movies per prompt. 

\begin{figure}%[h!]
    \centering
    \includegraphics[width=0.95\columnwidth]{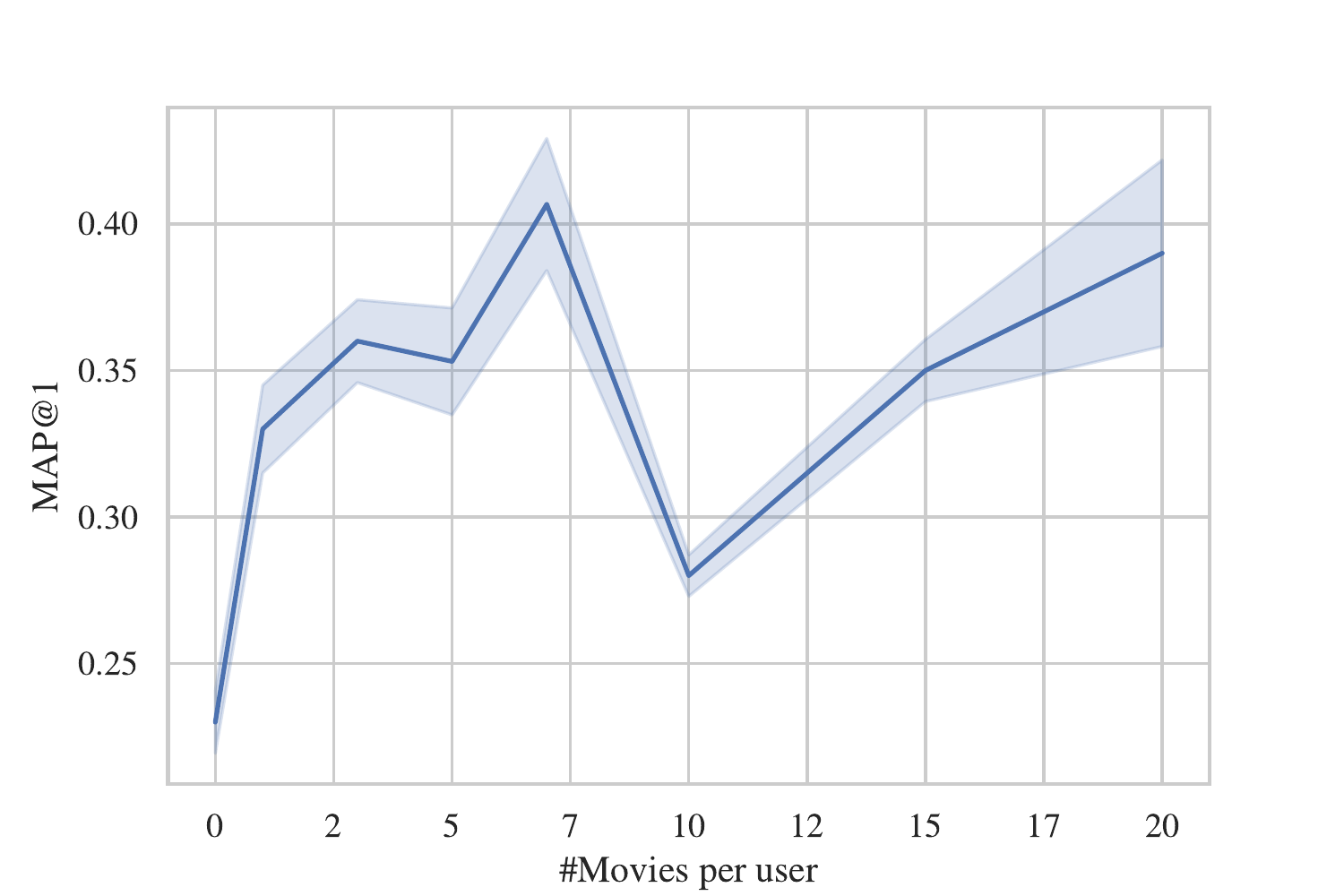}
    \caption{MAP@1 of LM models with a varying number of movies per user sampled in the input prompt.}
    \label{ratingsperuser}
\end{figure}

Figure \ref{ratingsperuser} shows that increasing the number of ratings per user has diminishing returns and lead to increasing instability, so specifying $n\approx 5$ seems to lead to the best results with the least user input. After 5 items, adding more items might make the prompt less natural, even though the LM seems to adapt when the number of items keeps increasing. It is also interesting to note that when we use an empty prompt, accuracy is above chance level because the LM captures some information about movie popularity.

\subsection{Comparison with matrix factorization and NSP}

We now use a matrix factorization as a baseline, with the Bayesian Personalised Ranking algorithm (BPR) \cite{rendle2012bpr}. Users and items are mapped to $d$ randomly initialized latent factors, and their dot product is used as a relevance score trained with ranking loss. We use \cite{salah2020cornac} implementation with default hyperparameters\footnote{\url{https://cornac.readthedocs.io/en/latest/models.html\#bayesian-personalized-ranking-bpr}, we experimented with other hyperparameter configurations but did not observe significant changes.} $d=10$ and a learning rate of $0.001$.

We also compare GPT-2 LM to BERT next sentence prediction \cite{Penha20} which models affinity scores with $\widehat{R}_{u,i}=\text{BERT}_{\text{NSP}}(p_u, {<}m_i{>})$, where $p_u$ is a prompt containing movies liked by $u$.
BERT was pretrained with contiguous sentence prediction task \cite{devlin-etal-2019-bert} and Penha et al. \cite{Penha20} proposed to use it as a way to probe BERT for recommendation capabilities.

\begin{figure}%[h!]
    \centering
    \includegraphics[width=0.95\columnwidth]{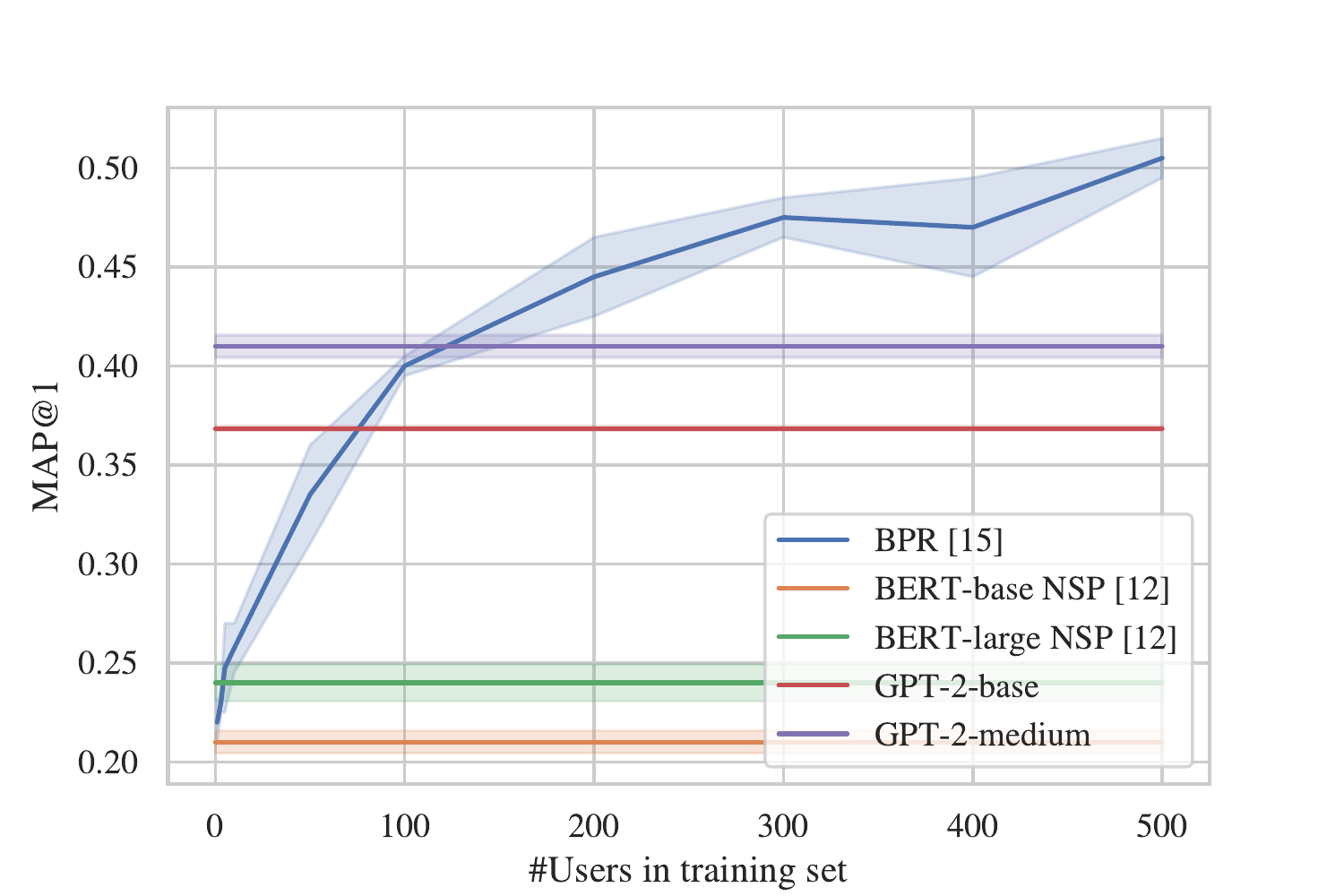}
    \caption{MAP@1 for BPR models with increasing numbers of users compared the zero-shot language models (with 0 training user). BERT-base and BERT-large respectively have 110M and 340M parameters. GPT-2-base and GPT-2-medium have 117M and 345M parameters. 
    }
    \label{baseline}
\end{figure}

Figure \ref{baseline} shows that the proposed LM-based recommendation significantly outperforms $\text{BERT}_{\text{NSP}}$. We explain the difference by the fact that LM are generative and actually model text likelihood, while next sentence prediction is discriminative and can be based on simple discursive coherence features.
It is also interesting to note that LM-based models outperform matrix factorization when there are few users, i.e $<50$ and $<100$ for \textsc{base} and \textsc{medium} GPT-2, which demonstrates that LM-based recommendation is viable for cold start regimes. Using models larger than the \textsc{base} versions lead to better results, however when we evaluated with larger versions (we did not perform the full experiments due to memory limitations), we did not see additional improvement, which could be explained by overfitting.

\subsection{Qualitative analysis of generations}
 \vspace{-0.1cm}

Up until there, we have used LM to score the likelihood of sequences. LM can also be used directly for text generation, unlike BERT. We here show LM-generated prompt completions randomly sampled in our dataset, using greedy decoding.

\paragraph{\textbf{Prompt (P1):}}
\textit{Forrest Gump, Blade Runner, Modern Times, Amelie, Lord of the Rings The Return of the King, Shaun of the Dead, Alexander, Pan's Labyrinth, Cashback, Avatar:}
\vspace{-.3cm}
\paragraph{\textbf{Completion (C1):}}
\textit{3, The Hunger Games: Mockingjay Part 2, King Arthur, A Feast for Crows, The Hunger Games: Catching Fire, Jackass, Jackass 2, King Arthur}

%\vspace{-.3cm}
\paragraph{\textbf{Prompt (P2):}}
\textit{Independence Day, Winnie the Pooh and the Blustery Day, Raiders of the Lost Ark, Star Wars Episode VI - Return of the Jedi, Quiet Man, Game, Labyrinth, Return to Oz, Song of the South, Matrix:}
\vspace{-.3cm}
\paragraph{\textbf{Completion (C2):}}
\textit{and many more. The list can be read by clicking on the relevant section at the left of the image. To access the list of releases}\\
 \vspace{-0.1cm}

Some prompts, i.e. (P1) generate valid movie names, but others, like (P2), do not. LM-based recommender do need a post-processing to match movie names in the possible sampled generations.

\section{Conclusion}
We showed that standard language models can be used to perform item recommendations without any adaptation and that they are competitive with supervised matrix factorization  when the number of users is very low (less than 100 users). LM can therefore be used to kickstart recommender systems if items are frequently discussed in the training corpora.
Further research could explore ways to adjust LM for recommendation purposes or to combine LM with matrix factorization into hybrid systems.
Another way to use of our findings would be to generate movie recommendation datasets by mining web data which could feed standard supervised recommendation techniques.

\section{Acknowledgements}
This work is part of the CALCULUS project, which is
funded by the ERC Advanced Grant H2020-ERC-2017­
ADG 788506\footnote{\url{https://calculus-project.eu/}}.

\bibliographystyle{splncs04}

\bibliography{references}

\clearpage

\end{document}